\pdfoutput=1

\documentclass[final]{cvpr}

\usepackage{times}
\usepackage{epsfig}
\usepackage{graphicx}
\usepackage{amsmath}
\usepackage{amssymb}

\usepackage{subcaption}
\usepackage{caption}
\usepackage{nidanfloat}
\usepackage{booktabs}
\usepackage{textgreek}
\usepackage{multirow}

\usepackage{subcaption}
\usepackage{enumitem}
\usepackage{mwe}
\usepackage{siunitx}

\usepackage{xspace}
\usepackage[dvipsnames]{xcolor}

\newcommand{\model}{$\pi$-GAN\xspace}
\newcommand{\siren}{{\sc siren}}

\newcommand{\pose}{\xi}
\newcommand{\coord}{\mathbf{x}}
\newcommand{\noise}{\mathbf{z}}
\newcommand{\dir}{\mathbf{d}}
\newcommand{\density}{\sigma}
\newcommand{\col}{\mathbf{c}}

\newcommand{\generatorparams}{\theta_G}
\newcommand{\generator}{G_{\generatorparams}}
\newcommand{\discriminatorparams}{\theta_D}
\newcommand{\discriminator}{D_{\discriminatorparams}}

\newcommand{\filmmult}{\boldsymbol{\gamma}}
\newcommand{\filmadd}{\boldsymbol{\beta}}

\newcommand{\implicitlayer}{\phi}
\newcommand{\implicit}{\Phi}

\newcommand{\nonlinearlayer}{\sin}

\newcommand{\myparagraph}[1]{\vspace{-8pt}\paragraph{#1}}

\usepackage[pagebackref=true,breaklinks=true,colorlinks,bookmarks=false]{hyperref}
\usepackage{cleveref}

\def\cvprPaperID{3471} 
\def\confYear{CVPR 2021}

\newcommand{\beginsupplement}{%
        \setcounter{section}{0}
        \renewcommand{\thesection}{\Alph{section}}
     }

\begin{document}


\title{pi-GAN: Periodic Implicit Generative Adversarial Networks for 3D-Aware Image Synthesis}




\author{
Eric R. Chan\thanks{These authors contributed equally to this work. Project page:\newline\url{https://marcoamonteiro.github.io/pi-GAN-website/}} $\,\,\,$
Marco Monteiro\footnotemark[1] $\,\,\,$
Petr Kellnhofer $\,\,\,$
Jiajun Wu $\,\,\,$
Gordon Wetzstein
\\
{Stanford University} \\
{\tt\small \{erchan,pkellnho,jiajunw,gordon.wetzstein\}@stanford.edu, monteiro.marcoa@gmail.com}
}

\maketitle
\thispagestyle{empty}
\global\csname @topnum\endcsname 0
\global\csname @botnum\endcsname 0

\begin{abstract}
We have witnessed rapid progress on 3D-aware image synthesis, leveraging recent advances in generative visual models and neural rendering. Existing approaches however fall short in two ways: first, they may lack an underlying 3D representation or rely on view-inconsistent rendering, hence synthesizing images that are not multi-view consistent; second, they often depend upon representation network architectures that are not expressive enough, and their results thus lack in image quality. We propose a novel generative model, named Periodic Implicit Generative Adversarial Networks (\model or pi-GAN), for high-quality 3D-aware image synthesis. \model leverages neural representations with periodic activation functions and volumetric rendering to represent scenes as view-consistent radiance fields. 
The proposed approach obtains state-of-the-art results for 3D-aware image synthesis with multiple real and synthetic datasets.
%
%
\end{abstract}

\begin{figure}
    \centering
    \includegraphics[width=\linewidth]{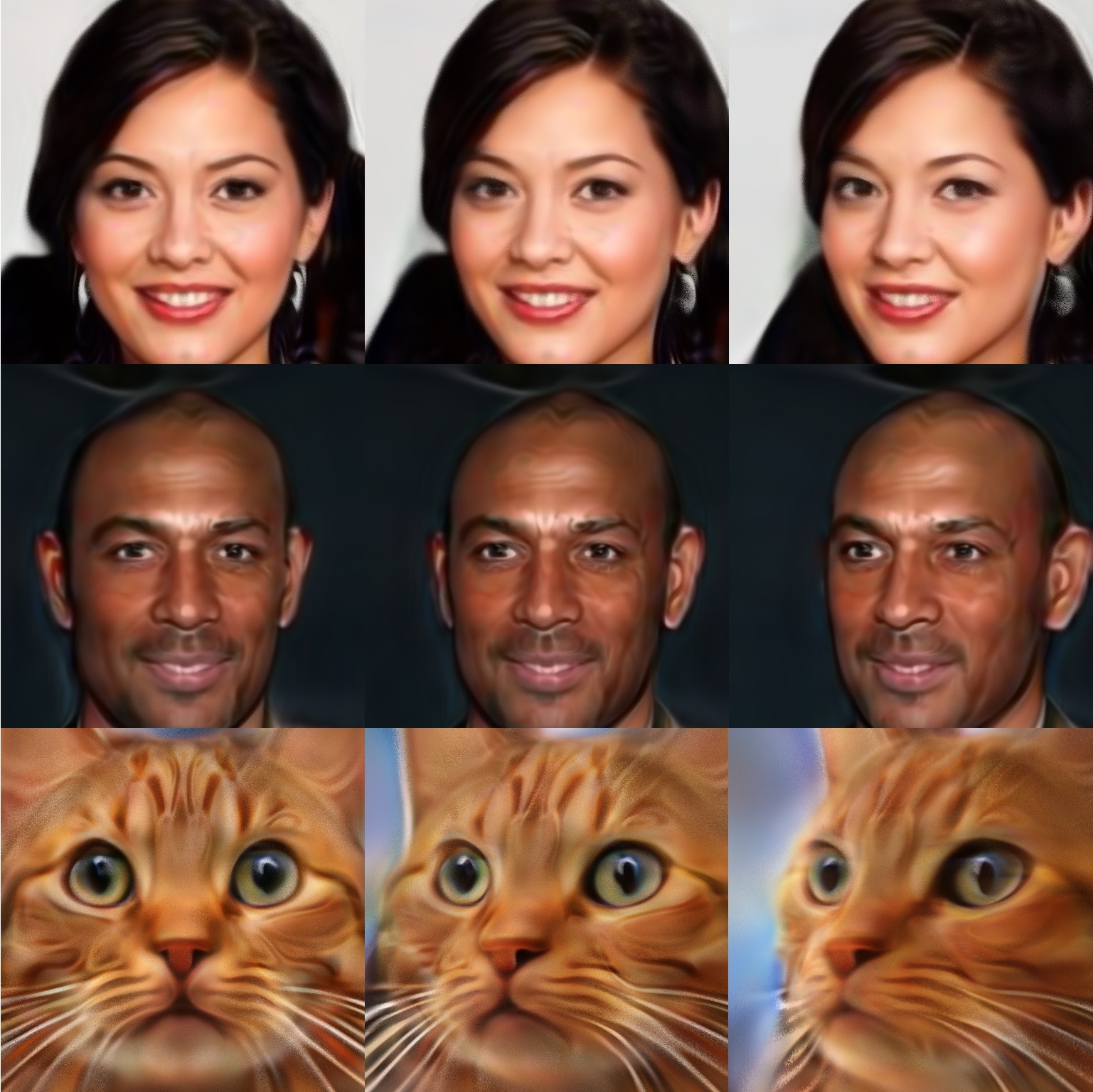}
    \caption{Selected examples synthesized by \model with CelebA~\cite{liu2015faceattributes} and Cats~\cite{cats} datasets.}
    \label{fig:teaser}
\end{figure}

\section{Introduction}
\label{sec:introduction}




Generative Adversarial Networks (GANs) are capable of generating high-resolution, photorealistic images~\cite{karras2018progressive,karras2019style,Karras2020stylegan2}. However, these GANs are often confined to two dimensions because of a lack of photorealistic 3D training data; therefore, they cannot support tasks such as synthesizing multiple views of a single object. 3D-aware image synthesis offers to learn neural scene representations unsupervised from 2D images. The learned representations can be used to render view-consistent images from new camera poses~\cite{hologan, graf, henzler2019platonicgan}.



Current solutions have achieved impressive results in decoupling identity from structure, allowing for the rendering of a single instance from multiple poses. Nevertheless, these approaches either lack multi-view consistency or fine detail.
Voxel-based approaches~\cite{henzler2019platonicgan} generate interpretable, true 3D representations, but are limited by computational complexity to low resolutions and coarse detail.
Convolutional approaches with deep-voxel representations~\cite{hologan, nguyen2020blockgan} take advantage of recent progress in convolutional GANs and can create finely detailed images.
However, because of their reliance on learned black-box rendering, these approaches fail to guarantee multi-view consistency and cannot easily generalize beyond the training distribution of camera poses at inference.
Recent approaches that leverage neural implicit representations~\cite{graf} incorporate representations based on neural network--parameterized radiance fields that ensure multi-view consistency and explicit camera control. Nonetheless, the implicit representations used by these approaches have so far been unable to effectively express fine details, leading to compromised image quality. 








We propose Periodic Implicit Generative Adversarial Networks (\model), a generative adversarial approach to unsupervised 3D representation learning from images. Given input noise, \model conditions an implicit radiance field represented by a \siren\ network~\cite{sitzmann2020siren}, a fully-connected network with periodic activation functions. The conditioned radiance field maps a 3D location and 2D viewing direction to a view-dependent radiance and view-independent volume density~\cite{kajiya1984ray,Max:1995}. Using a differentiable volume rendering approach that relies on classical volume rendering techniques, we can render the radiance field from arbitrary camera poses~\cite{mildenhall2020nerf}.




\model improves upon the image quality and view-consistency of previous approaches to 3D-aware image synthesis, as shown in Figure~\ref{fig:teaser}. The proposed method utilizes a 
\siren-based neural radiance field representation to encourage multi-view consistency, allowing rendering from a wide range of camera poses and providing an interpretable 3D structure. The \siren\ implicit scene representation, which makes use of periodic activation functions, is more capable than ReLU implicit representations at representing fine details and enables \model to render sharper images than previous works.




Beyond introducing \model, we make two additional technical contributions.
First, we observe that while existing work has conditioned ReLU-based radiance fields through concatenation of the input noise to one or more layers, conditioning-by-concatenation is sub-optimal for implicit neural representations with period activations (\siren s). We instead propose to use a mapping network to condition layers in the \siren\ through feature-wise linear modulation (FiLM)~\cite{perez2017film,dumoulin2018feature-wise}. This contribution can more generally be applied to \siren\ architectures beyond GANs. Second, we introduce a progressive growing strategy, inspired by previous successes in 2D convolutional GANs~\cite{karras2018progressive}, to accelerate training and offset the increased computational complexity of 3D GANs.


We obtain state-of-the-art 3D-aware image synthesis results on real-world and synthetic datasets, demonstrate that our method generalizes to new viewpoints, and has applications to novel view synthesis. Moreover, the 5D spatio-angular radiance field representation used by \model allows for an interpretable 3D proxy shape to be extracted via the marching cubes algorithm~\cite{lorensen1987marching}. While these proxy shapes may not be as high quality as those estimated by single-view shape reconstruction methods tailored to this task~\cite{wu2020unsupervised}, they often end up resulting in a fair approximation, all without explicit supervision.

Our contributions in this paper include the following:
\begin{itemize}[noitemsep]
    
    \item We introduce \siren-based implicit GANs as a viable alternative to convolution GAN architectures.
    \vspace{3pt}\item We propose a mapping network with FiLM conditioning and a progressive growing discriminator as key components to achieve high quality results with our novel \siren-based implicit GAN. 
    \vspace{3pt}\item We demonstrate view consistency and explicit camera control as advantages of approaches that rely on an underlying neural radiance field representation and classical rendering.
    \vspace{3pt}\item We achieve state-of-the-art results on 3D-aware image synthesis from unsupervised 2D data on the CelebA~\cite{liu2015faceattributes}, Cats~\cite{cats}, and CARLA~\cite{DBLP:journals/corr/abs-1711-03938, graf} datasets.
\end{itemize}




\section{Related Work}
\label{sec:related}

\paragraph{Neural representations and rendering.}
Emerging neural implicit scene representations promise 3D-structure-aware, continuous, memory-efficient representations for parts~\cite{genova2019learning,genova2019deep}, objects~\cite{park2019deepsdf,michalkiewicz2019implicit,atzmon2019sal,gropp2020implicit,yariv2020multiview,davies2020overfit,chabra2020deep}, or scenes~\cite{eslami2018neural,sitzmann2019srns,jiang2020local,peng2020convolutional,sitzmann2020siren}. These can be supervised with 3D data, such as point clouds, and optimized as either signed distance functions~\cite{park2019deepsdf,michalkiewicz2019implicit,atzmon2019sal,gropp2020implicit,sitzmann2019srns,jiang2020local,peng2020convolutional,sitzmann2019metasdf,kellnhofer:2021} or occupancy networks~\cite{mescheder2019occupancy,chen2019learning}. 
Using neural rendering~\cite{tewari2020state}, implicit neural representations can also be trained using multiview 2D images~\cite{saito2019pifu,sitzmann2019srns,Oechsle2019ICCV,Niemeyer2020CVPR,mildenhall2020nerf,yariv2020multiview,liu2020neural,jiang2020sdfdiff,liu2020dist}.
Temporally aware extensions ~\cite{Niemeyer2019ICCV} and multimodal variants with part-level semantic segmentation~\cite{kohli2020inferring} have also been proposed. 

Among these approaches, sinusoidal representation networks (\siren)~\cite{sitzmann2020siren} and neural radiance fields (NeRF)~\cite{mildenhall2020nerf} are most closely related to our work. Specifically, we use \siren{} as the representation network architecture of our framework combined with a neural rendering technique inspired by NeRF. Both \siren{} and NeRF, however, have only been explored in the context of overfitting to individual objects or scenes, whereas we study the combination of aspects of these seminal works for applications in 3D GANs. Exploring the unique challenges of training a neural implicit GAN supervised by natural 2D data is one of the core contributions of our work.

\myparagraph{Generative 3D-aware image synthesis.}
Generative Adversarial Nets (GANs)~\cite{goodfellow2014generative}, or more generally the paradigm of adversarial learning, have led to significant progress in various image synthesis tasks, including image generation~\cite{radford2016unsupervised,karras2018progressive,karras2019style,Karras2020stylegan2}, image-to-image translation \cite{zhu2017unpaired}, interactive image editing~\cite{wang2018high}, and learning from partial and noisy observations~\cite{bora2018ambientgan}. These methods operate on the 2D space of pixels, ignoring the 3D nature of our physical world. This has limited the application of these generative models in tasks such as view synthesis.

Visual Object Networks~\cite{zhu2018visual} and PrGANs~\cite{Gadelha:2017} learn to synthesize 2D images by first generating a voxelized 3D shape using a 3D-GAN~\cite{wu2016learning} and then projecting it into 2D.
HoloGAN~\cite{hologan} and BlockGAN~\cite{nguyen2020blockgan} have extended the system by incorporating a volumetric but implicit 3D representation. While these methods attempt to model the 3D structure of the object in the synthesized image, the use of an explicit volume representation has constrained their resolution~\cite{lunz2020inverse}. 
Szab\'o~\etal~\cite{Szabo:2019} and Liao~\etal~\cite{Liao2020CVPR} instead proposed to model 3D shapes as meshes and collections of primitives for image synthesis, respectively. However, these representations lack the expressiveness needed to synthesize high-fidelity pictures.


The work most similar to ours is GRAF~\cite{graf}, which learns a generative model for implicit radiance fields for 3D-aware image synthesis. Although \model operates in a similar setting, its network architecture and training strategy differ from GRAF in several ways. First, we use \siren\ rather than a positionally encoded ReLU MLP as a choice of neural implicit representation. Second, GRAF conditioned its MLP generator on both a shape noise code and an appearance noise code by concatenation; in contrast, we leverage a StyleGAN-inspired mapping network, which conditions the entire MLP on a single input noise vector through FiLM conditioning. Third, we utilize a progressive growing strategy during training. Finally, we did not employ a patch-based discriminator, as used by GRAF, as \siren{} is prone to local overfitting to the last batch if sufficient coverage of the space is not maintained. Our experiments demonstrate that all of our innovations are critical to high-quality image synthesis results.
    
Beyond unconditional 3D-aware image generation, there is an orthogonal line of work on conditional reconstruction of 3D shape and texture from partial observations. These reconstructions can later be used for novel view synthesis. Various 3D representations have been considered for the task, including voxels~\cite{henzler2019platonicgan,tulsiani2017learning}, meshes~\cite{kanazawa2018learning,chen2019learning,goel2020shape,henderson2020leveraging,mustikovela2020self}, point clouds~\cite{tatarchenko2016multi}, a depth map~\cite{wu2020unsupervised}, and implicit functions~\cite{rajeswar2020pix2shape,tulsiani2020implicit}. Some of these methods are also grounded in adversarial training. While these methods focus on 3D reconstruction, \model aims to learn an unconditional generative model of radiance fields.

\section{Methods}
\label{sec:methods}


\model is a generative approach to learning radiance field representations from unlabeled 2D images, with the goal of synthesizing high-quality view consistent images. 
Traditional 2D GANs, such as StyleGAN~\cite{karras2019style}, take in a latent vector $\noise \sim p_\noise$ and directly produce a 2D image. Instead of directly generating a 2D image from the input noise, $\noise$, our generator $\generator(\noise, \pose)$ produces an implicit radiance field conditioned on $\noise$. This radiance field is rendered using volume rendering to produce a 2D image from some camera pose $\pose$.

At training time, the generated images are directed to a traditional convolutional discriminator for adversarial training. At test time, the radiance field can be rendered from arbitrary camera poses to produce view-consistent images.




\begin{figure*}[ht]
    \centering
    \begin{subfigure}[t]{1.0\textwidth}
        \includegraphics[width=\textwidth]{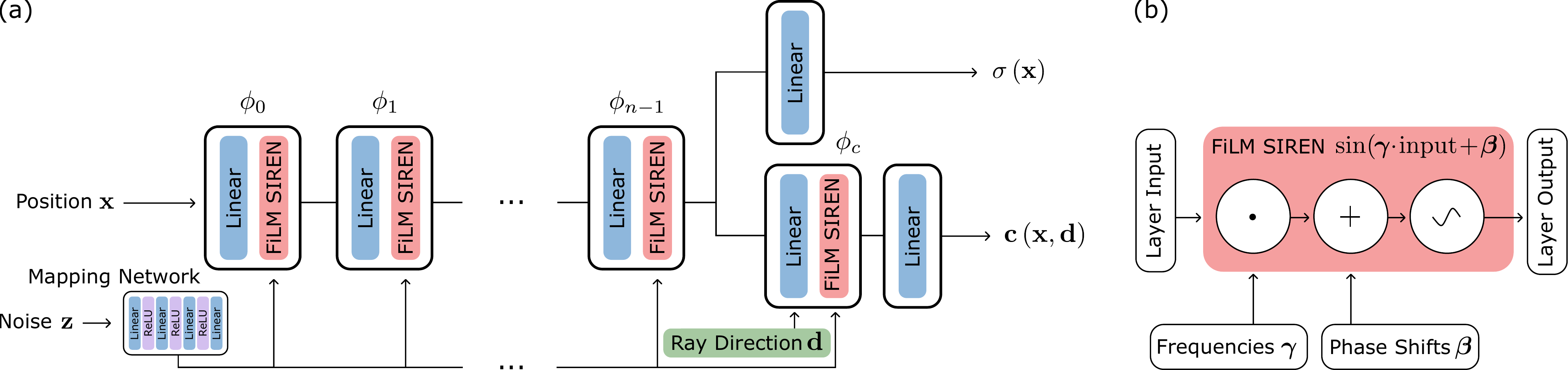}
        \phantomcaption
        \label{fig:network_architecture_diagram}
    \end{subfigure}
    \begin{subfigure}[t]{0.4\textwidth}
        \phantomcaption
        \label{fig:filmsiren_diagram}
    \end{subfigure}
    \caption{The \model generator architecture.}
\end{figure*}

\subsection{\siren-Based Implicit Radiance Field}


We represent 3D objects implicitly with a neural radiance field, which is parameterized as a multilayer perceptron (MLP) that takes as input a 3D coordinate in space $\coord = (x,y,z)$ and the viewing direction $\dir$. The neural radiance field outputs both the spatially varying density $\density(\coord) : \mathbb{R}^3 \rightarrow \mathbb{R}$ and the view-dependent color $(r,g,b) = \col(\coord,\dir) : \mathbb{R}^5 \rightarrow \mathbb{R}^3$. Moreover, we leverage a StyleGAN-inspired mapping network to condition the \siren{} on a noise vector $\noise$ through FiLM conditioning~\cite{perez2017film,dumoulin2018feature-wise}.



As shown in Figure~\ref{fig:network_architecture_diagram}, we formalize the FiLM-ed \siren{} backbone of our representation as
%
\begin{align}
    \implicit \left( \mathbf{x} \right) = &  \implicitlayer_{n-1} \circ \implicitlayer_{n-2} \circ \ldots \circ \implicitlayer_0  \left( \mathbf{x} \right),  \\
    & \implicitlayer_i \left( \mathbf{x}_i \right)  = \nonlinearlayer \left( \filmmult_i \cdot \left( \mathbf{W}_i \mathbf{x}_i + \mathbf{b}_i \right) + \filmadd_i \right),
\end{align}
where $\implicitlayer_i: \mathbb{R}^{M_i} \mapsto \mathbb{R}^{N_i}$ is the $i^{th}$ layer of an MLP. It consists of an affine transform defined by the weight matrix $\mathbf{W}_i \in \mathbb{R}^{N_i \times M_i}$ and the biases $\mathbf{b}_i\in  \mathbb{R}^{N_i}$ applied on the input $\coord_i\in\mathbb{R}^{M_i}$, followed by the sine nonlinearity applied to each component of the resulting vector (Figure~\ref{fig:filmsiren_diagram}). Our mapping network is a simple ReLU MLP, which takes as input a noise vector $\noise$ and outputs the frequencies $\filmmult_i$ and phase shifts $\filmadd_i$, which condition each layer of the \siren. 


We found this mapping network to be more expressive than concatenation-based conditioning. It yielded image-quality improvements, both for conditioning ReLU-based and \siren-based neural implicit representations. The ablation studies shown in Sec.~\ref{sec:experiments:ablations} give further insight into these conditioning methods.

Both density and color of our implicit volume are then defined as
\begin{align}
    \density \left( \coord \right) & = \mathbf{W}_{\sigma} \implicit \left( \mathbf{x} \right)  + \mathbf{b}_{\sigma}, \\
    \col \left( \coord, \dir \right) & = \mathbf{W}_{c}  \implicitlayer_{c} \left( \left[ \implicit \left( \mathbf{x} \right), \dir \right]^T \right) + \mathbf{b}_{c},
\end{align}
where $\mathbf{W}_{\sigma / c}$ and $\mathbf{b}_{\sigma / c}$ are additional weight and bias parameters.





\subsection{Neural Rendering}

\begin{figure}
    \centering
    \includegraphics[width=0.7\columnwidth]{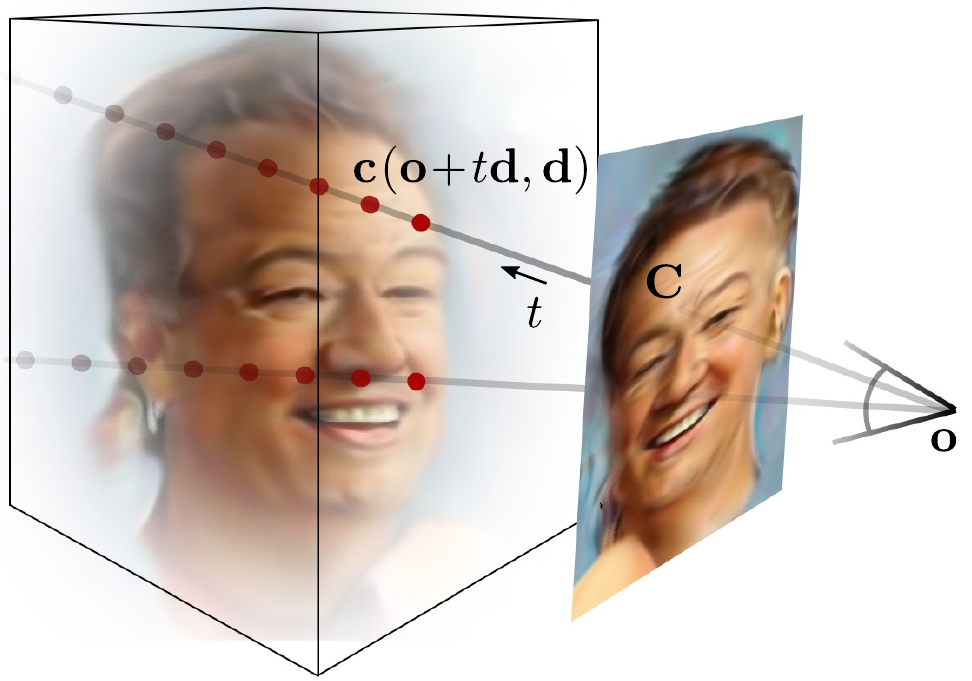}
    \caption{A visualization of our neural volume rendering procedure. Given a conditioned radiance field, we cast rays from the camera origin $\mathbf{o}$, sample density \textsigma\ and color $\mathbf{c}$ values along each ray, and calculate pixel color $\mathbf{C}$ using Eq.~\ref{eqn:volumetric_rendering}.}
    \label{fig:volume_rendering}
\end{figure}

We render a neural radiance field from arbitrary camera poses $\pose$ using neural volume rendering. For this purpose, we employ a pinhole camera model and cast rays from the camera origin $\mathbf{o}$ to compute the integrals along each ray through the volume. At every sample, our generator predicts the volume density $\density$ and color $\col$. The pixel color $\mathbf{C}$ for a camera ray $\mathbf{r}(t) = \mathbf{o} + t \dir$ with near and far bounds $t_n$ and $t_f$ is then calculated using the volume rendering equation~\cite{Max:1995}:
\begin{align}
\label{eqn:volumetric_rendering}
\begin{split}
    \mathbf{C} (\mathbf{r}) & = \int_{t_n}^{t_f} T(t) \density \left(\mathbf{r}(t) \right) \col \left(\mathbf{r}(t),\dir \right)dt,
\\
& \text{where} \quad
T(t)  = \text{exp} \left(-\int_{t_n}^{t}\sigma(\mathbf{r}(s))ds \right).
\end{split}
\end{align}


Our approach implements a discretized form of this equation using the stratified and hierarchical sampling approach introduced by NeRF~\cite{mildenhall2020nerf} (see Figure~\ref{fig:volume_rendering}).

This neural rendering approach, which is also adopted by GRAF~\cite{graf}, has several advantages over previous 3D-to-2D projections. Neural rendering allows for explicit control over camera pose, focal length, aspect ratio, and other parameters, while simple projections, such as those used by HoloGAN~\cite{hologan}, are restricted to representing poses in the training dataset. 




\subsection{Discriminator}
Following ProgressiveGAN~\cite{karras2018progressive}, we use a convolutional discriminator $\discriminator$ with parameters $\discriminatorparams$ that grows progressively. We begin training at low resolutions and high batch sizes, during which the generator can focus on producing coarse shapes. As training progresses, we increase the image resolution and add new layers to the discriminator to handle the higher resolutions and discriminate fine details. For most experiments, we begin training at $32 \times 32$ and double the resolution twice during training, up to $128 \times 128$. In practice, we found this progressive growing strategy to allow for larger batch sizes at the beginning of training, which helped to stabilize and speed training (see Sec.~\ref{sec:experiments:ablations}). Final results are rendered by sampling $512 \times 512$ pixels.

Unlike ProgressiveGAN~\cite{karras2018progressive}, our generator architecture does not grow; instead, we increase the resolution of the generator by sampling rays more densely from the same implicit representation.

\subsection{Training Details}
At training time, we randomly sample camera poses $\pose$ from a distribution $p_\pose$. The pose distributions for each dataset are known a priori and approximated as either Gaussian, for CelebA and Cats, or uniform, for CARLA (see supplement for details). In our experiments, we constrained camera positions to the surface of a unit sphere and directed the camera to point towards the origin. At training time, pitch and yaw along the sphere were sampled from a distribution that was tuned according to the dataset. Real images $I$ are sampled from the training set with distribution $p_I$. We use the non-saturating GAN loss with R1 regularization~\cite{gan_convergence}:
%
\begin{align}
\label{eqn:eqlabel}
\begin{split}
\mathcal{L}(\theta, \phi) & = \mathbf{E}_{\noise\sim p_z, \xi\sim p_\xi}[f(\discriminator(\generator(\noise, \pose)))]
\\
& +\mathbf{E}_{I\sim p_\mathcal{D}}[f(-\discriminator(I)) + \lambda \lvert \nabla \discriminator (I) \rvert ^ 2],
\\
& \text{where} \quad f(u) = -\log(1 + \exp(-u)).
\end{split}
\end{align}


We train \model in a generative adversarial framework in which a generator and discriminator compete in a zero sum game. Our generator tries to minimize Equation~\ref{eqn:eqlabel}, while the discriminator simultaneously tries to maximize Equation~\ref{eqn:eqlabel}. We use the Adam optimizer with $\beta_1 = 0$, $\beta_2 = 0.9$. We initialize learning rates to \SI{5e-5} for the generator and \SI{4e-4} for the discriminator, decayed over training to \SI{1e-5} and \SI{1e-4} respectively. Further training and implementation details can be found in the supplemental materials.

\section{Experiments and Analysis}
\label{sec:experiments}


  

In this section, we first evaluate the quality of images generated by \model. We then demonstrate that it learns 3D representations that enables synthesizing images at unseen poses. We also include ablation studies to justify our use of sinusoidal activations and mapping network conditioning.

\subsection{Evaluating Image Quality}

\paragraph{Datasets.} We evaluate \model on the real-world CelebA~\cite{liu2015faceattributes} and Cats~\cite{cats} datasets, as well as the synthetic CARLA~\cite{DBLP:journals/corr/abs-1711-03938, graf} dataset. CelebA contains 200,000 high-resolution face images of 10,000 different celebrities. We crop the images from the top of the hair to the bottom of the chin. The Cats dataset contains 6,444 $128 \times 128$ images of cat heads. The CARLA dataset contains 10k images of 16 car models with random texture and color properties, rendered with the Carla Driving simulator. We train and evaluate at $128 \times 128$ resolution for all datasets and models. We evaluate all models using a moving average of parameters.

\myparagraph{Baselines.}
We compare against two previous approaches to 3D-aware image synthesis: HoloGAN~\cite{hologan} and Generative Radiance Fields (GRAF)~\cite{graf}. Baseline models were obtained as pre-trained checkpoints directly from the authors or trained until convergence using the recommended hyperparameters.

\myparagraph{Qualitative results.}
\begin{figure*}
    \centering
    \includegraphics[width=\textwidth]{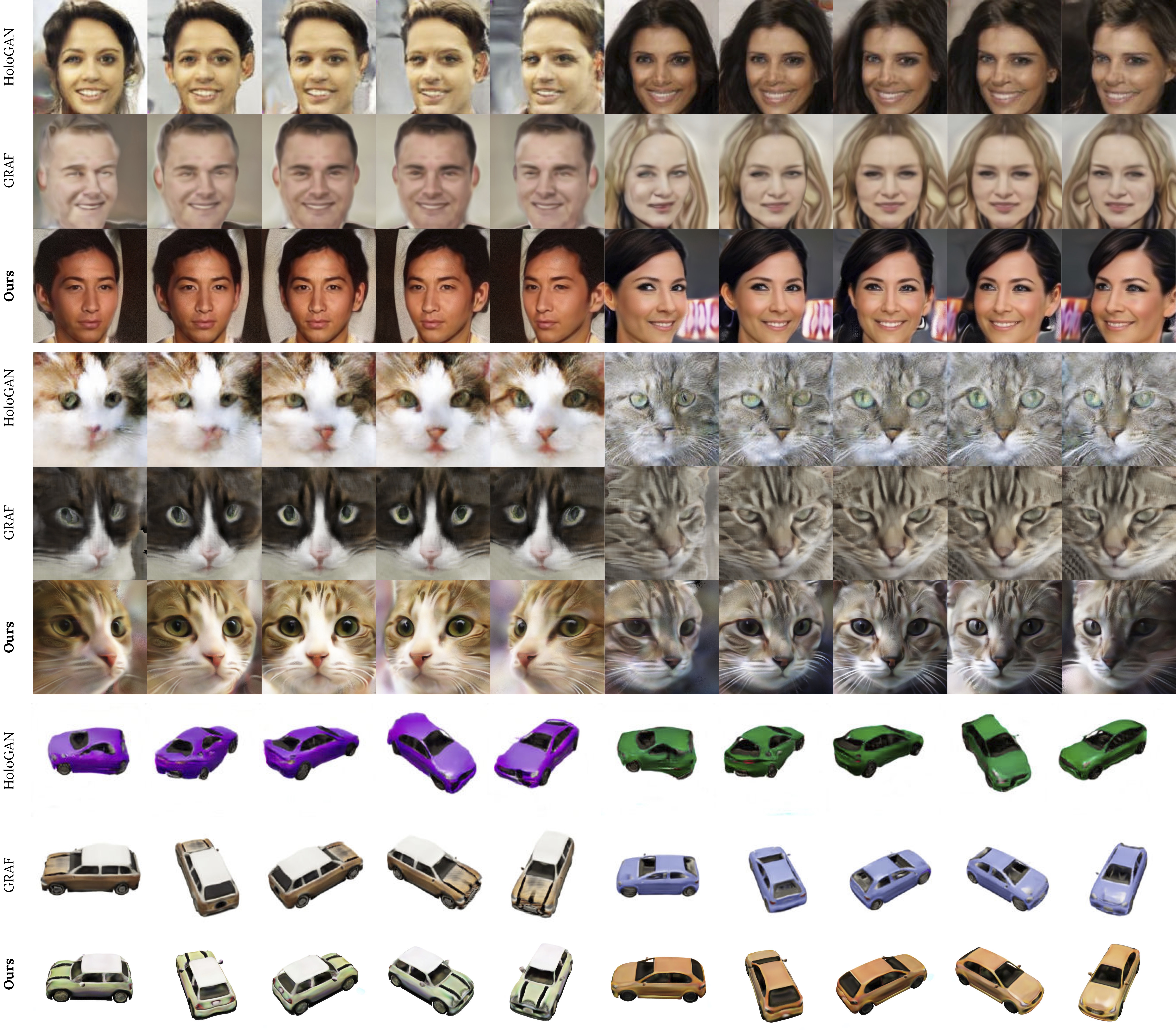}
    \caption{Qualitative comparison on CelebA, Cats, and CARLA.} 
    \label{fig:qualitative_comparison}
\end{figure*}

\begin{figure*}
    \centering
    \includegraphics[width=\textwidth]{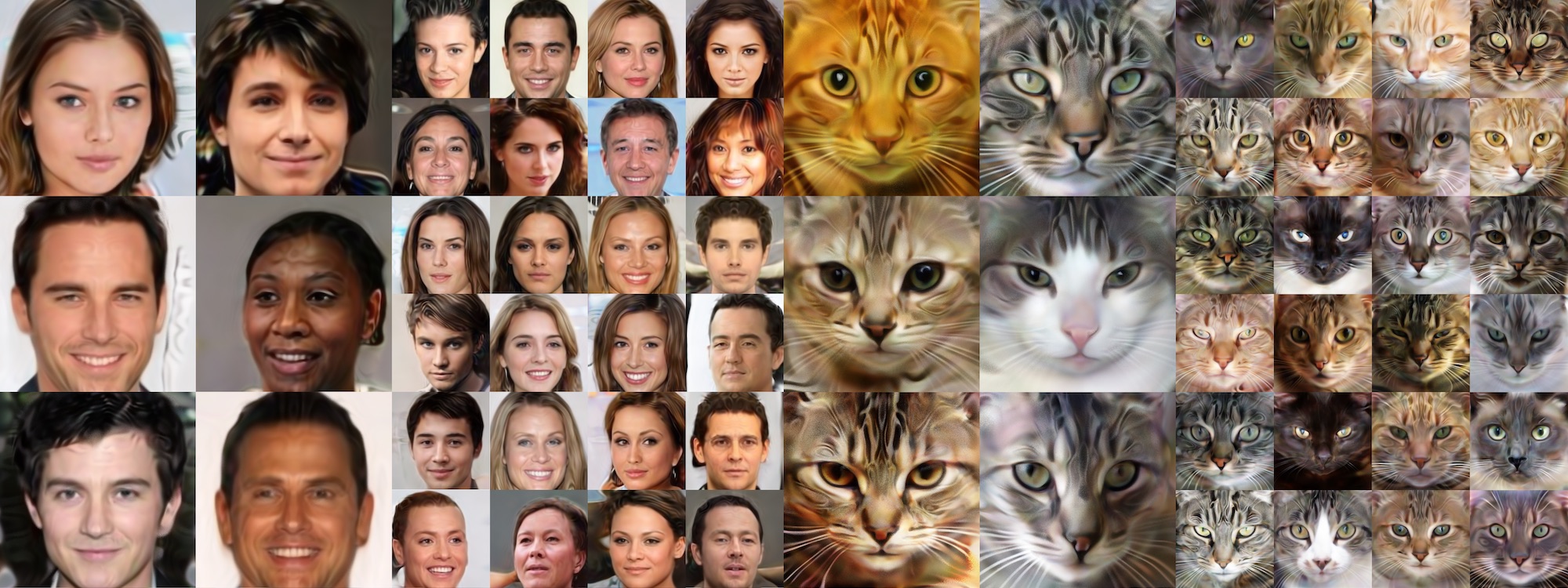}
    \caption{Uncurated generated faces, corresponding to the first 30 random seeds.}
    \label{fig:uncurated_celeba}
\end{figure*}

Figure~\ref{fig:qualitative_comparison} compares images generated by \model, HoloGAN, and GRAF on three datasets.

Qualitatively, HoloGAN achieves good image quality but suffers from multi-view inconsistency. Although it generally produces sharp images, identity shift is visible across rotations, particularly at the edges of the training distribution. HoloGAN struggled on the synthetic CARLA dataset, which featured much larger variations in viewpoint than CelebA or Cats. Previous papers were also unable to obtain consistent HoloGAN baselines on this dataset~\cite{graf}.

GRAF, which allows for explicit camera control, is more capable than HoloGAN at recovering wide viewing angles. Because it utilizes a 3D representation, it renders different views of the same scene with less identity shift than HoloGAN. However, GRAF is less capable than HoloGAN at rendering fine details such as hair and teeth, and generally produces images that are more cartoon-ish and less lifelike than HoloGAN.

Our \model combines fine details with the ability to represent a wide range of camera angles. Compared with HoloGAN and GRAF, it better recreates details such as individual teeth (CelebA) and whiskers (Cats). Because we represent each instance with a radiance field, 
\model generates images that are inherently view consistent, have minimal identity shift, and that recover a wide range of angles. 


\myparagraph{Quantitative results.}
We evaluate image quality using Frechet Inception Distance (FID)~\cite{DBLP:journals/corr/HeuselRUNKH17}, Kernel Inception Distance (KID)~\cite{binkowski2018demystifying}, and Inception Score~\cite{DBLP:journals/corr/SalimansGZCRC16}. Tables \ref{tbl:CelebA}, \ref{tbl:Cats}, and \ref{tbl:Carla} show a quantitative comparison on CelebA, Cats, and CARLA, respectively. We show significant improvements in image quality metrics compared with baselines, particularly on real-world datasets with fine details. Additional results, including precision-recall plots \cite{precision_recall_distributions}, are provided in the supplemental material.

Our evaluation was consistently performed across all models for Table~\ref{tbl:ImageQuality}. Note that specific experiment parameters, such as image crop, may differ from those used by other authors.

\begin{table*}[ht]
\centering
 \begin{subfigure}[b]{0.3\textwidth}
    \caption{CelebA @ 128 $\times$ 128}
    \label{tbl:CelebA}
    
\begin{tabular}{llll}
\toprule
                 & FID $\downarrow$ & KID $\downarrow$  & IS $\uparrow$  \\ 
\midrule
HoloGAN          & 39.7  & 2.91 & 1.89   \\
GRAF             & 41.1  & 2.29    & 2.34   \\
\model            & \textbf{14.7}   & \textbf{0.39}    & \textbf{2.62}   \\
\bottomrule
\end{tabular}

 \end{subfigure}
 \hfill
 \begin{subfigure}[b]{0.3\textwidth}
    \caption{Cats @ 128 $\times$ 128}
    \label{tbl:Cats}
    
    \begin{tabular}{llll}
    \toprule
                     & FID $\downarrow$ & KID $\downarrow$  & IS $\uparrow$  \\ 
    \midrule
    HoloGAN          & 40.4 & 3.30 & 2.03   \\
    GRAF             & 28.9 & 1.43 & 1.66   \\
    \model             & \textbf{16.8}   & \textbf{0.92}  & \textbf{2.06}   \\
    \bottomrule
    \end{tabular}
    
 \end{subfigure}
 \hfill
 \begin{subfigure}[b]{0.3\textwidth}
    \caption{CARLA @ 128 $\times$ 128}
    \label{tbl:Carla}
    \begin{tabular}{llll}
    \toprule
                     & FID $\downarrow$ & KID $\downarrow$  & IS $\uparrow$  \\ 
    \midrule
    HoloGAN          & 67.5 & 3.95 & 3.52   \\
    GRAF             & 41.7  &  2.43  & 3.70   \\
    \model             & \textbf{29.2}   & 
    \textbf{1.36}    &  \textbf{4.27}   \\
    \bottomrule
    \end{tabular}
    
 \end{subfigure}
    \caption{FID, KID mean$\times$100, and IS for CelebA, Cats, and CARLA datasets.}
    \label{tbl:ImageQuality}
\end{table*}

\subsection{Generating Approximate 3D Representations}

A key advantage of our approach over previous CNN attempts at 3D representation learning is that by generating an 
implicit radiance field, our model learns an underlying 3D-structure-aware representation. This representation allows for explicit camera control, naturally lends itself to rendering poses that were uncommon or unseen at training time, and is interpretable.





\myparagraph{Extrapolation to rare or unseen camera poses.}

\model relies on an underlying 3D structural representation and offers explicit camera control. Like previous methods that offer explicit camera control (e.g.,~\cite{graf}), it more readily renders views and poses outside of the training dataset distribution than previous methods that rely on black-box representations or projections (e.g.,~\cite{hologan}). 

\begin{figure}[t]
    \centering
    \includegraphics[width=\linewidth]{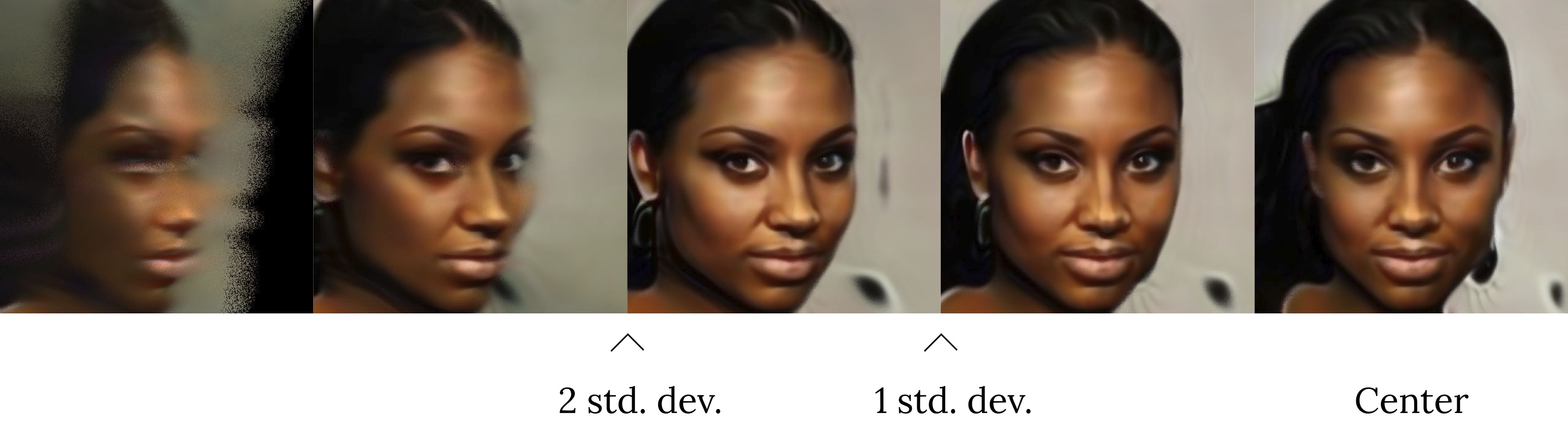}
    \caption{\model is capable of rendering views from steep angles, producing reasonable results even beyond two standard deviations of camera yaw on CelebA. Face yaw on CelebA is approximately zero-centered Gaussian, with a standard deviation of 17º from the centerline.}
    \label{fig:high_angles}
\end{figure}

Figure~\ref{fig:high_angles} shows that the explicit camera control and representation naturally generalizes to rendering views even from steep angles, although visual artifacts are stronger at the edges of the camera distribution. This is a consequence of the distribution of CelebA images being imbalanced towards front-facing images. As shown in Figure \ref{fig:qualitative_comparison}, CARLA, which features uniformly distributed poses, did not suffer from this issue.

\begin{figure}[t]
    \centering
    \includegraphics[width=\linewidth]{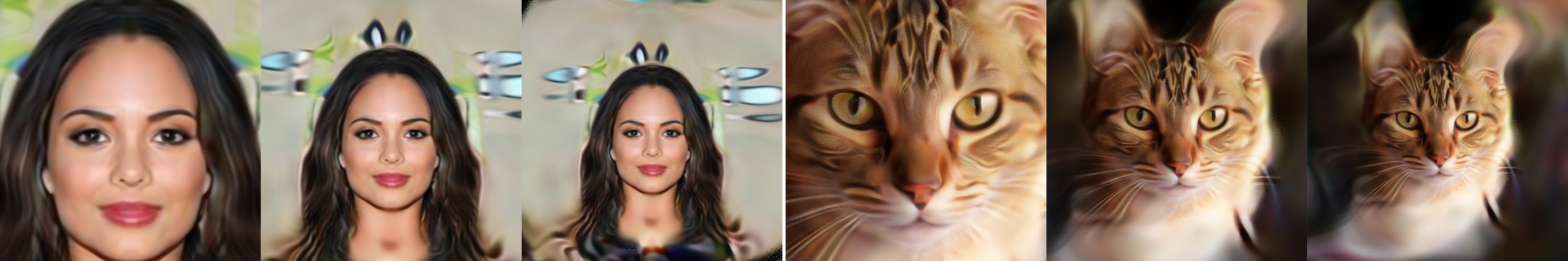}
    \caption{Explicit camera control at inference enables rendering views completely absent from the training distribution of camera poses. Although \model was trained only on close-up images, it extrapolates to zoomed-out poses.}
    \label{fig:zoom_out}
\end{figure}

Figure~\ref{fig:zoom_out} illustrates that, despite only training on tightly cropped images, the radiance field extrapolates when we zoom out the camera. Because the radiance field may be rendered from any of a wide variety of angles at training time, the generator is encouraged to produce a radiance field that represents the entire scene, even if only a small portion will be visible in any single image.

\begin{figure}[t]
    \centering
    \includegraphics[width=\linewidth]{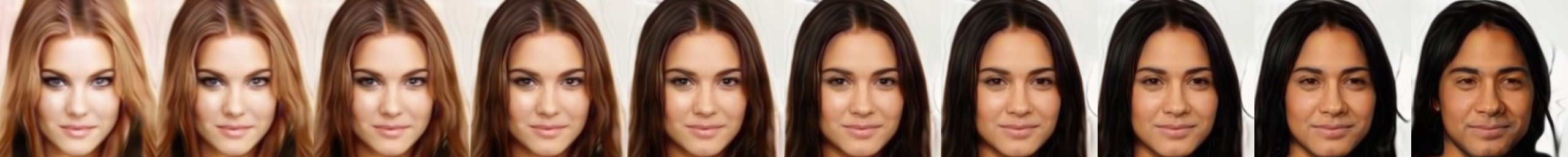}
    \caption{Linearly interpolating between two latent codes.}
    \label{fig:interpolation}
\end{figure}

To demonstrate that the latent space learned by \model is semantically meaningful, we show the results of interpolating between two latent codes in Figure~\ref{fig:interpolation}.

\myparagraph{Interpreting the 3D representation.}

\begin{figure}[t]
    \centering
    \includegraphics[width=\linewidth]{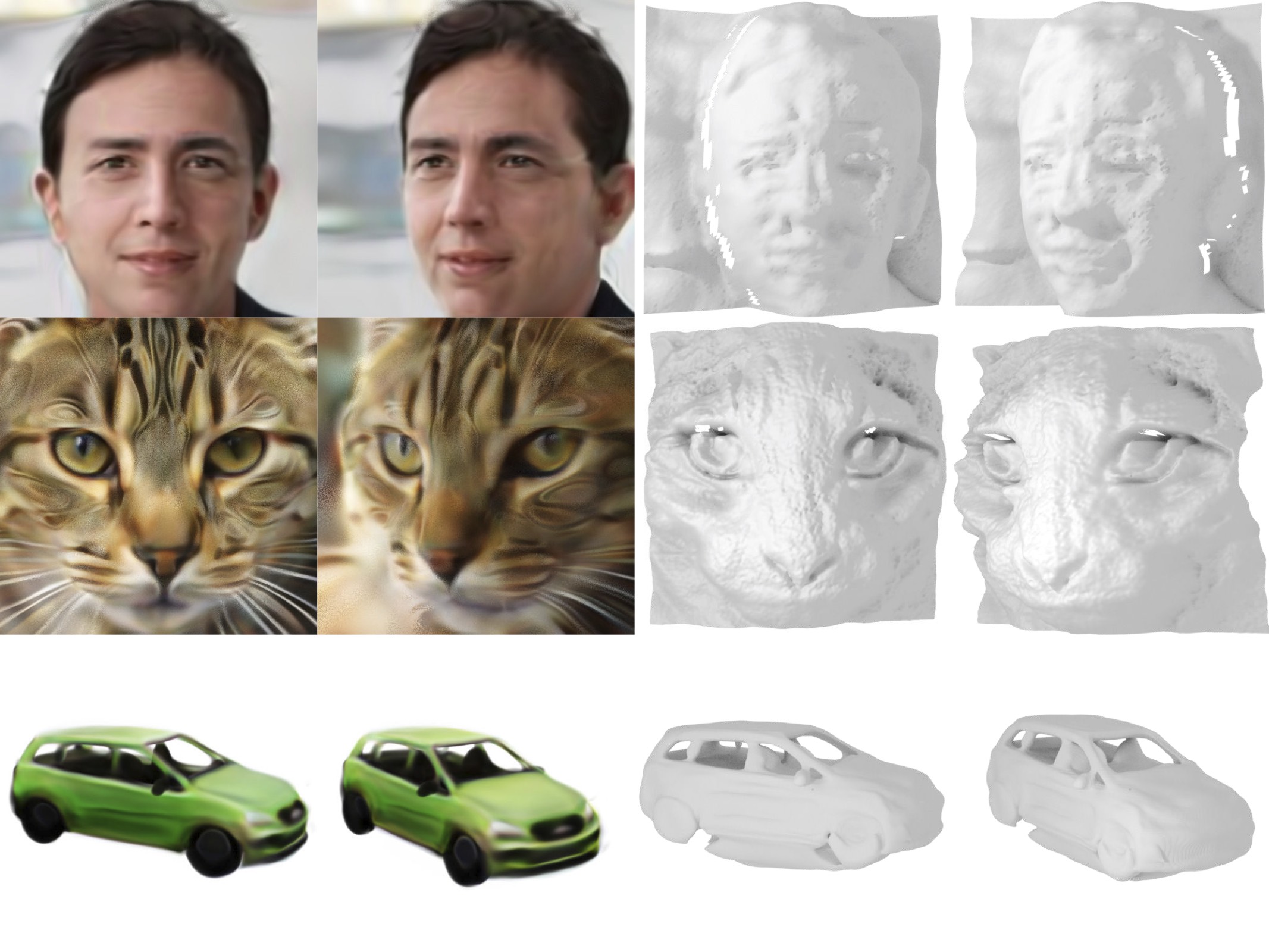}
    \caption{We can extract a proxy 3D representation as a mesh, either by projecting a depth-map (CelebA, Cats), or through marching cubes (CARLA).}
    \label{fig:marching_cubes}
\end{figure}

Although the color output of the implicit representation depends on ray direction to allow for view-dependent effects, such as specularities, the density output $\density$ is completely view independent, resulting in a view-consistent 3D structure that represents a proxy shape of the scene. This 3D structure can be extracted and visualized using the marching cubes algorithm~\cite{lorensen1987marching} on the density output of the conditioned radiance field to produce a surface mesh. Figure~\ref{fig:marching_cubes} shows 3D models extracted from the 3D representation.

\subsection{Ablations}
\label{sec:experiments:ablations}

   \begin{table}
       \centering
         \begin{tabular}{lcc}
         \toprule
            \multirow{2}{*}{Conditioning} &  \multicolumn{2}{c}{Architecture}  \\ 
         \cmidrule{2-3}
            & ReLU P.E. & Sine \\
         \midrule
         Concatenation       &   32.0  & 21.6  \\
         Mapping Network                     &   26.8  & \textbf{5.15}   \\
         \bottomrule
         \end{tabular}
       \caption{FID scores on CelebA @ $64 \times 64$, when comparing network architectures with different activation functions and conditioning methods.}
       \label{tbl:ablations}
   \end{table}



We ablate sinusoidal activations and mapping network conditioning to better understand their individual contributions. We compare radiance fields with sinusoidal activations against radiance fields with ReLU activations and positional encodings (P.E.)~\cite{mildenhall2020nerf}. Moreover, we evaluate radiance fields conditioned with a mapping network and FiLM conditioning against radiance fields conditioned via concatenation~\cite{graf}. Table~\ref{tbl:ablations} summarizes the results of these experiments. Ablations were conducted at $64 \times 64$ in order to save computational resources. Sinusoidal activations and mapping network conditioning each yielded improvements against their respective baselines. However, the combined model, with both sinusoidal activations and a mapping network, was more effective than the sum of its parts.

  \begin{figure}
    \centering
    \includegraphics[width=1.0\linewidth]{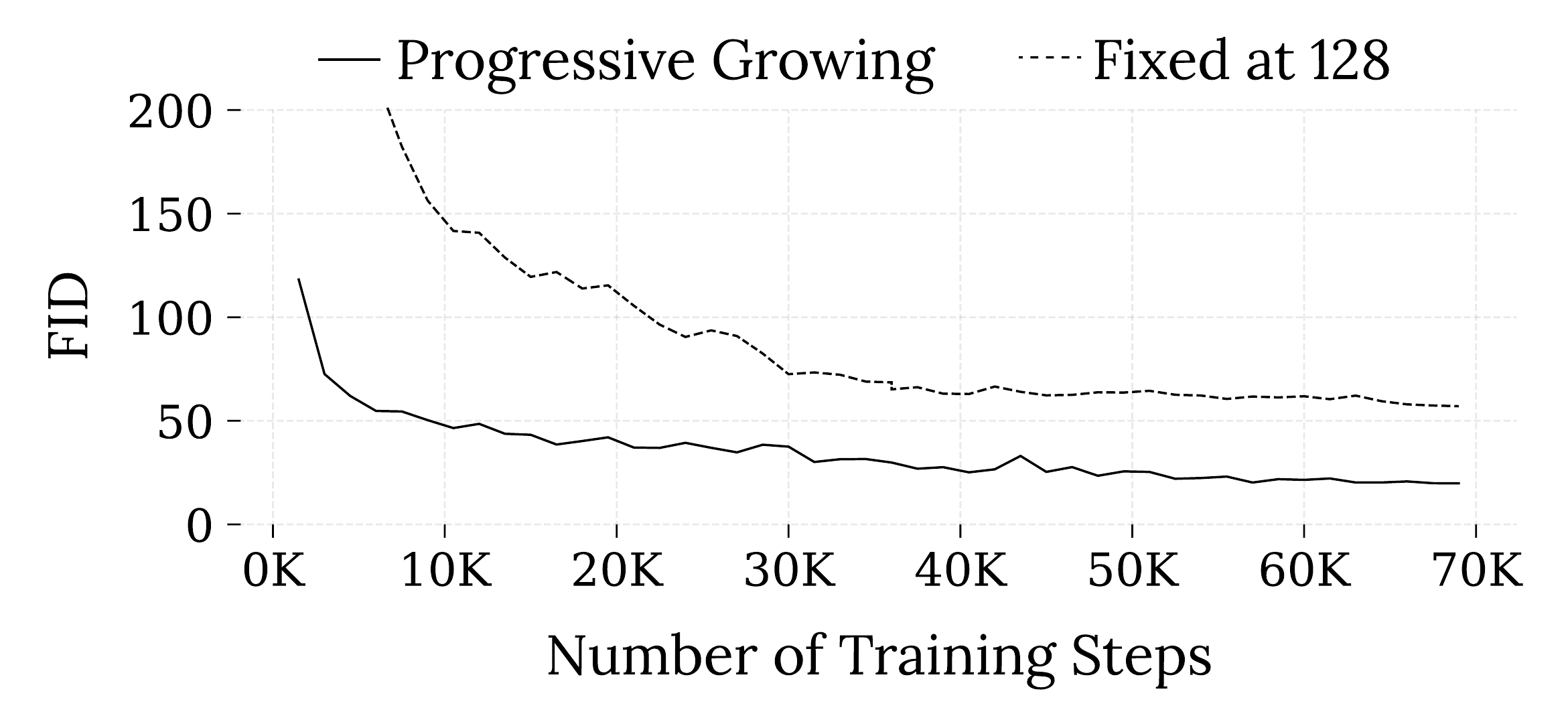}
    \caption{Ablation study for training \model with and without progressive growing on CelebA @ $128 \times 128$}
    \label{fig:progressive_ablation}
\end{figure}

Figure~\ref{fig:progressive_ablation} compares early training steps for a model trained with progressive growing against a model initialized to the full $128 \times 128$ image resolution. Because computational complexity grows quadratically with image size, progressive growing, which begins at low resolutions, allows for the use of much larger batch sizes at the start of training. The large batch sizes are helpful in stabilizing training, while also allowing for a higher throughput in images per iteration. As others have found before us~\cite{karras2018progressive}, progressive growing, and the larger batch sizes it enables, helped ensure quality and diversity for generated images.

\section{Discussion}
\label{sec:discussion}
\paragraph{Applications to novel view synthesis.}

\begin{figure}[t]
    \centering
    \includegraphics[width=\linewidth]{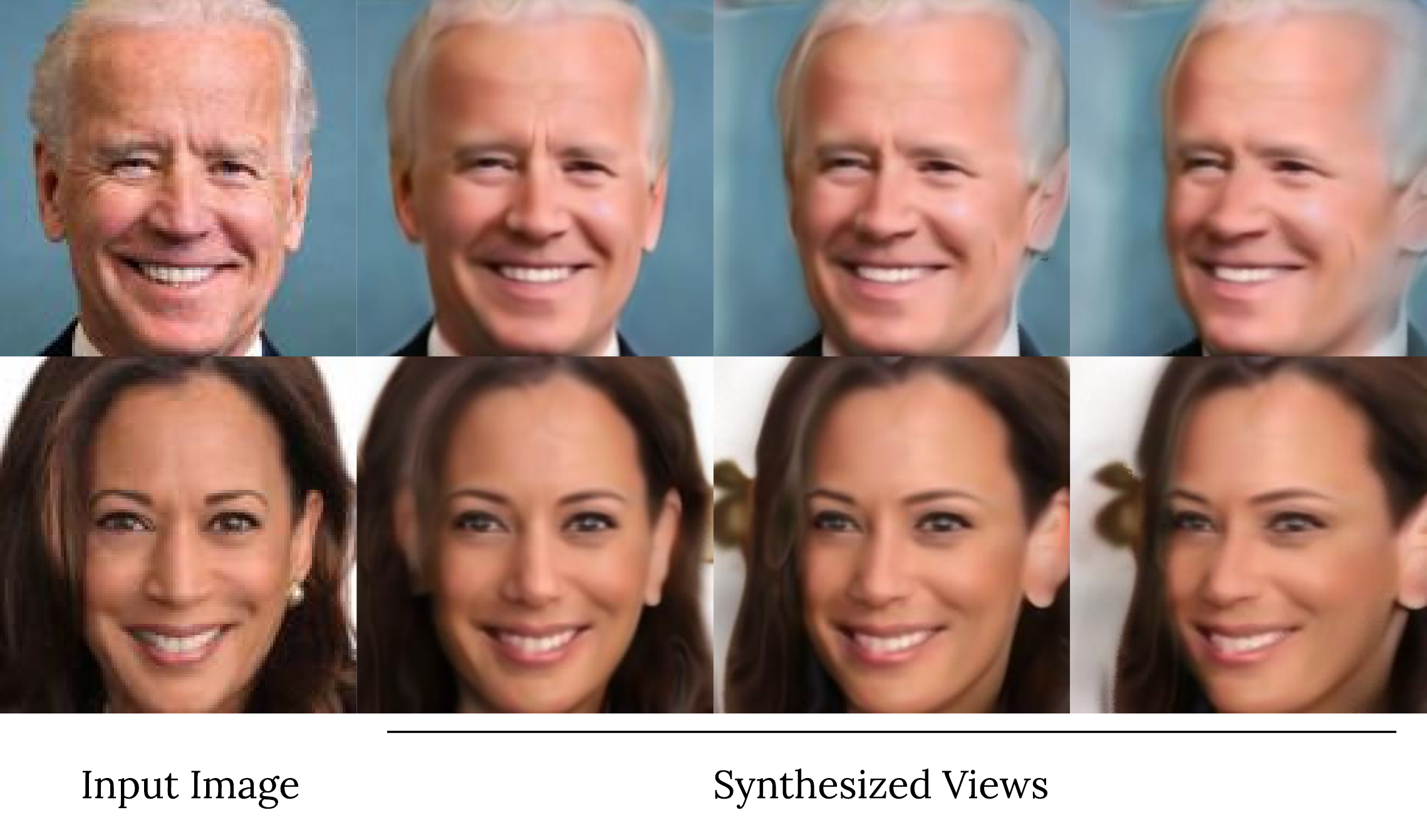}
    \caption{Using a trained \model generator, we can optimize a radiance field to fit an input image and synthesize novel views from arbitrary camera poses.}
    \label{fig:inverse_render}
\end{figure}

Figure~\ref{fig:inverse_render} demonstrates that it is possible to use a trained generator, without modifications, to perform single-view reconstruction using the procedure described by Karras et al.~\cite{Karras2020stylegan2}. For this purpose, we freeze the parameters of our implicit representation and seek the frequencies $\filmmult_i$ and phase shifts $\filmadd_i$ for each MLP layer $i$ which produce a radiance field that, when rendered, best matches the target image. Additional details are found in the supplement.



\myparagraph{Failure modes, limitations, and future work.}

While \model has demonstrated considerable improvements to image quality for 3D-aware image synthesis, there remain a plethora of avenues for future work.

Although the unsupervised learning of 3D shapes was not the focus of this work, \model nevertheless produces interpretable and view-consistent 3D representations that capture the 3D structures of objects. Future work could focus on refining the quality of extracted meshes, with \model as a viable solution to learning shapes from unposed images.

\begin{figure}
    \centering
    \includegraphics[width=\linewidth]{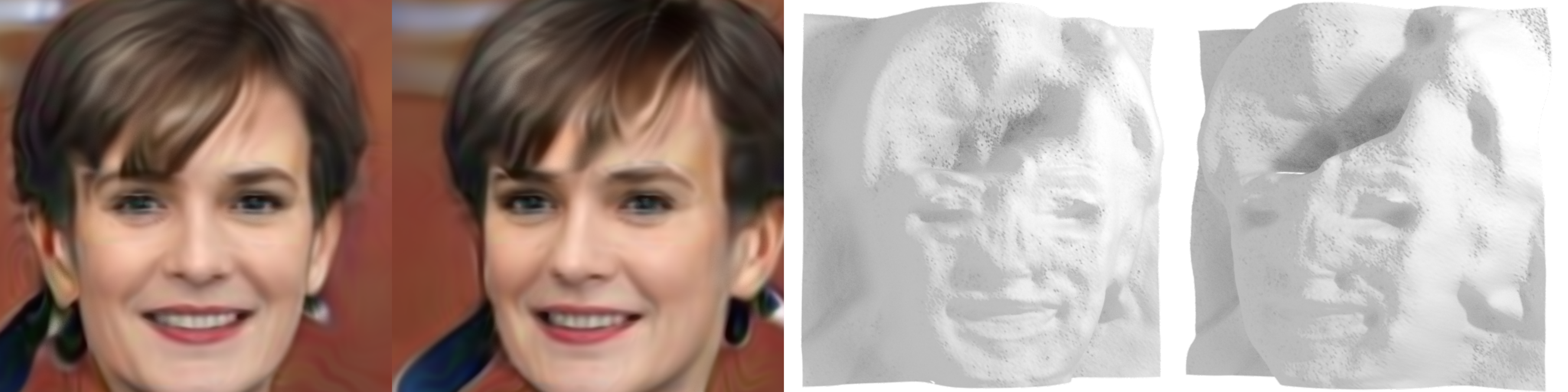}
    \caption{In a failure case reminiscent of the hollow-face illusion, our model sometimes generates objects with inverted sections.}
    \label{fig:hollow_ambiguity}
\end{figure}

In certain cases, \model can generate a radiance field that creates viable images when rendered from each direction but nonetheless fails to conform to the 3D shape that we would expect. As Figure \ref{fig:hollow_ambiguity} demonstrates, a concave face is a valid geometric solution, given the constrained range of poses the discriminator sees at training. Further investigation may reveal insights that could resolve such ambiguities.

While \model has made strides in improving image quality for 3D-aware image synthesis, much work remains before implicit GANs can match the image quality of state-of-the-art 2D-convolutional GANs~\cite{Karras2020stylegan2,brock2018large,karras2018progressive}. Future work may produce solutions to remaining visual artifacts and further improve image quality.
\model is computationally expensive compared to traditional 2D GANs because the complexity of training the generator scales not only with image size but also with depth along each ray. More efficient render techniques could lower the computational barrier and allow for larger, sharper images.

\myparagraph{Ethical considerations.}
While our inverse rendering results only reconstruct static images, the method could be extended to generate fake photos or videos of real people (DeepFakes). DeepFakes pose a societal threat, and we do not condone using our work to generate fake images or videos of any person with the intent of spreading misinformation or tarnishing their reputation. We also recognize a lack of diversity in our faces results, stemming from the implicit bias in the CelebA dataset. 

\myparagraph{Conclusion.}

Photorealistic 3D-aware image synthesis has many exciting applications in vision and graphics. With our work, we take a significant step towards this goal.


\myparagraph{Acknowledgements.}
Thanks to Matthew Chan for fruitful discussions and to Stanford HAI for AWS Cloud Credits. J.W. was supported by the Samsung Global Research Award and Autodesk. G.W. was supported by an NSF CAREER Award (IIS 1553333) and a PECASE from the ARO. 

{\small
\bibliographystyle{ieee_fullname}
\bibliography{egbib}
}

\beginsupplement
\clearpage












\section{Novel View Synthesis Details}

We demonstrate a potential application of \model: we can use a trained generator, without modifications, to perform single-view reconstruction. We base our method on the inverse projection procedure outlined by Karras et al.~\cite{Karras2020stylegan2}.

We freeze the parameters of our implicit representation and seek the frequencies $\filmmult_i$ and phase shifts $\filmadd_i$ for each MLP layer $i$ which produce a radiance field that, when rendered, best matches the target image. We initialize $\filmmult_i$ and $\filmadd_i$ to $\bar{\filmmult_i}$ and $\bar{\filmadd_i}$, the center of mass of frequencies and phase shifts for each layer. We calculate $\bar{\filmmult_i}$ and $\bar{\filmadd_i}$ simply by averaging the frequencies and phase shifts of ten thousand random noise vector inputs. We then run gradient descent to minimize the mean-squared-error image reconstruction loss. We additionally introduce an $\mathcal{L}_2$ penalty with a weight of 0.1 during the optimization process to prevent $\filmmult_i$ and $\filmadd_i$ from straying too far from $\bar{\filmmult_i}$ and $\bar{\filmadd_i}$. We optimize the frequencies and phase shifts with the Adam optimizer over 700 iterations. We initialize the learning rate to 0.01, decaying by a factor of 0.5 every 200 iterations.


\section{Model Details}

\paragraph{Mapping Network.}
The mapping network is parameterized as an MLP with three hidden layers of 256 units each. The mapping network uses leaky-ReLU activations with a negative slope of 0.2.


\paragraph{\siren-based Implicit Radiance Field.}
The FiLMed-\siren~\cite{sitzmann2020siren} backbone of the generator is parameterized as an MLP with eight FiLMed-\siren\ hidden layers of 256 units each.

\paragraph{Discriminator.}

\begin{table}
\centering
\caption{Discriminator architecture, showing progressive growing stages.}
\label{tbl:discriminator_layers}
\begin{tabular}{lll} 
\toprule
\begin{tabular}[c]{@{}l@{}}\\\end{tabular}                                                                                            & Activation                                                                                  & Output Shape                                                                                             \\ 
\midrule
\begin{tabular}[c]{@{}l@{}}Input Image\\Adapter Block (1$\times$1)\\Coord Conv 1 (3$\times$3)\\Coord Conv 2 (3$\times$3)\\Avg Pool Downsample\end{tabular} & \begin{tabular}[c]{@{}l@{}}-\\LeakyReLU (0.2)\\LeakyReLU (0.2)\\LeakyReLU (0.2)\\-\\ \end{tabular} & \begin{tabular}[c]{@{}l@{}}3$\times$128$\times$128~\\64$\times$128$\times$128~\\128$\times$128$\times$128\\128$\times$128$\times$128~\\128$\times$64$\times$64\\\end{tabular}  \\ 
\midrule
\begin{tabular}[c]{@{}l@{}}Coord Conv 1 (3$\times$3)\\Coord Conv 2 (3$\times$3)\\Avg Pool Downsample\\\end{tabular}                                 & \begin{tabular}[c]{@{}l@{}}LeakyReLU (0.2)\\LeakyReLU (0.2)\\-\\\end{tabular}                  & \begin{tabular}[c]{@{}l@{}}256$\times$64$\times$64\\256$\times$64$\times$64\\256$\times$32$\times$32\\\end{tabular}                                \\ 
\midrule
\begin{tabular}[c]{@{}l@{}}Coord Conv 1 (3$\times$3)\\Coord Conv 2 (3$\times$3)\\Avg Pool Downsample\\\end{tabular}                                 & \begin{tabular}[c]{@{}l@{}}LeakyReLU (0.2)\\LeakyReLU (0.2)\\-\\\end{tabular}                  & \begin{tabular}[c]{@{}l@{}}400$\times$32$\times$32\\400$\times$32$\times$32\\400$\times$16$\times$16\\\end{tabular}                                \\ 
\midrule
\begin{tabular}[c]{@{}l@{}}Coord Conv 1 (3$\times$3)\\Coord Conv 2 (3x3)\\Avg Pool Downsample\\\end{tabular}                                 & \begin{tabular}[c]{@{}l@{}}LeakyReLU (0.2)\\LeakyReLU (0.2)\\-\\\end{tabular}                  & \begin{tabular}[c]{@{}l@{}}400$\times$16$\times$16\\400$\times$16$\times$16\\400$\times$8$\times$8\\\end{tabular}                                  \\ 
\midrule
\begin{tabular}[c]{@{}l@{}}Coord Conv 1 (3$\times$3)\\Coord Conv 2 (3$\times$3)\\Avg Pool Downsample\\\end{tabular}                                 & \begin{tabular}[c]{@{}l@{}}LeakyReLU (0.2)\\LeakyReLU (0.2)\\-\\\end{tabular}                  & \begin{tabular}[c]{@{}l@{}}400$\times$4$\times$4\\400$\times$4$\times$4\\400$\times$2$\times$2\\\end{tabular}                                      \\ 
\midrule
Conv 2d (2$\times$2)                                                                                                                         &                                                                                             & 1$\times$1$\times$1                                                                                                    \\
\bottomrule
\end{tabular}
\end{table}

Table~\ref{tbl:discriminator_layers} shows the architecture of the progressive discriminator. We begin training at low resolutions and progressively add discriminator stages while upsampling image size. In order to smooth transitions between upsamples, we fade in the contributions of new layers over ten-thousand iterations. We utilized CoordConv layers~\cite{liu2018intriguing} and residual connections~\cite{he2016deep} throughout the discriminator.We considered using a patch discriminator similar to GRAF, but found it leads to uneven image quality as SIREN is prone to local overfitting to the last batch if sufficient coverage of the space is not maintained. 

\begin{table}
   \centering
      \caption{FID, KID mean $\times$ 100, and IS for \model on CelebA, Cats, and CARLA datasets. }
        \label{tbl:piGAN64}
     \begin{tabular}{llll}
\toprule
                 & FID $\downarrow$ & KID $\downarrow$  & IS $\uparrow$  \\ 
\midrule
CelebA @ $64 \times 64$         & 5.15  & 0.09  & 2.28   \\
Cats @ $64 \times 64$           & 7.36  & 0.23    & 2.07   \\
CARLA  @ $64 \times 64$         &  13.59  &  0.34  &  3.85  \\
\bottomrule
\end{tabular}

\end{table}
\begin{figure}
    \centering
    \includegraphics[width=0.8\linewidth]{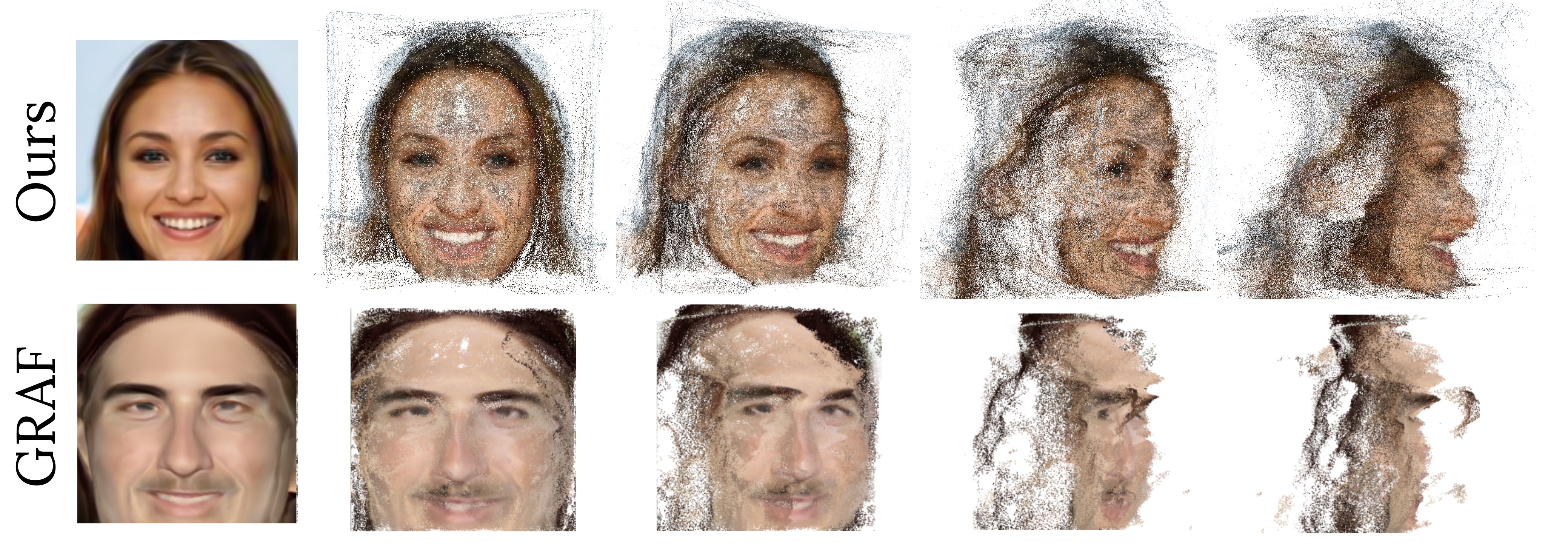}
    \caption{COLMAP reconstructions for models trained on CelebA, obtained by running COLMAP with default parameters and no known camera poses; GRAF's results were from their supplement.}
    \label{fig:COLMAP}
\end{figure}

\begin{figure}
    \centering
    \includegraphics[width=0.8\linewidth]{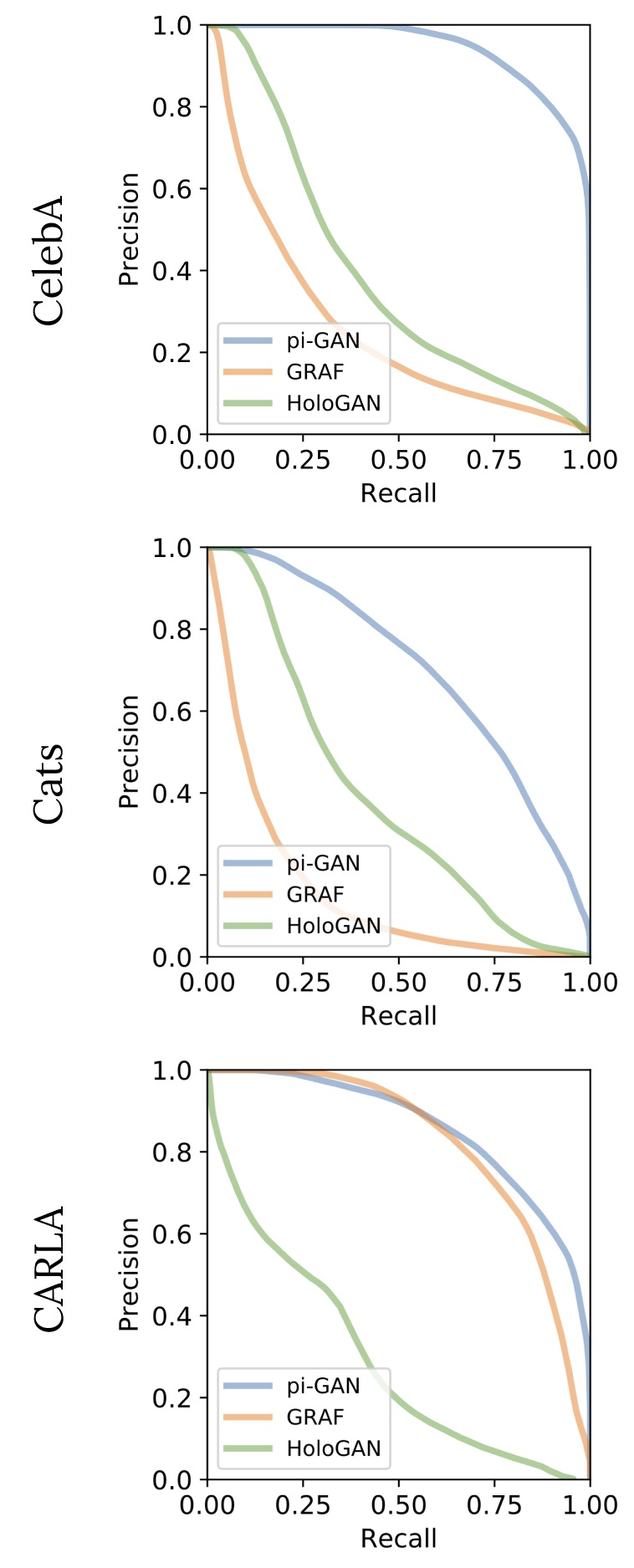}
    \caption{Precision-recall plots for \model, GRAF, and HoloGAN on CelebA, Cats, and CARLA.}
    \label{fig:Precision-Recall}
\end{figure}

\section{Additional Training Details}
We train the majority of our models across two RTX 6000 GPUs or a single RTX 8000 GPU. We begin training at a resolution of $32\times32$, with an initial batch size of 120. At each upsample, we drop the batch size by a factor of four to keep the models and generated images in memory. At higher resolutions, we aggregate across mini-batches to keep an effective batch size at or above 12, given our GPU constraints. To further reduce memory usage, we used PyTorch's Automatic Mixed Precision (AMP). \model trained for 10 hours at 32$\times$32, 10 hours at 64$\times$64, and 36 hours at 128$\times$128. Certain rendering and camera parameters were tuned according to the dataset. We use the true pose distribution when it is known, e.g. for synthetic datasets, otherwise we make a guess and tune the distribution as a hyperparameter. We sample camera poses for CelebA from a normal distribution, with a vertical standard deviation of 0.15 radians and a horizontal standard deviation of 0.3 radians. We sample camera poses for Cats from a uniform distribution, with horizontal range $(-0.75, 0.75)$ and vertical range $(-0.4, 0.4)$. We sample poses for CARLA uniformly from the upper hemisphere. We tune the number of samples along each ray to balance memory consumption and depth resolution. We use 24 samples per ray for CelebA and Cats and 64 samples per ray for CARLA. We utilize a pinhole perspective camera with a field of view of 12º for CelebA, 12º for Cats, and 30º for CARLA.

\section{\model results @ $64 \times 64$}

Table~\ref{tbl:piGAN64} includes additional quantitative results, evaluated at $64 \times 64$, in order to allow for comparisons of \model against models evaluated at lower resolutions.

\section{Additional Visual Results}
\begin{figure*}
    \centering
    \includegraphics[width=\textwidth]{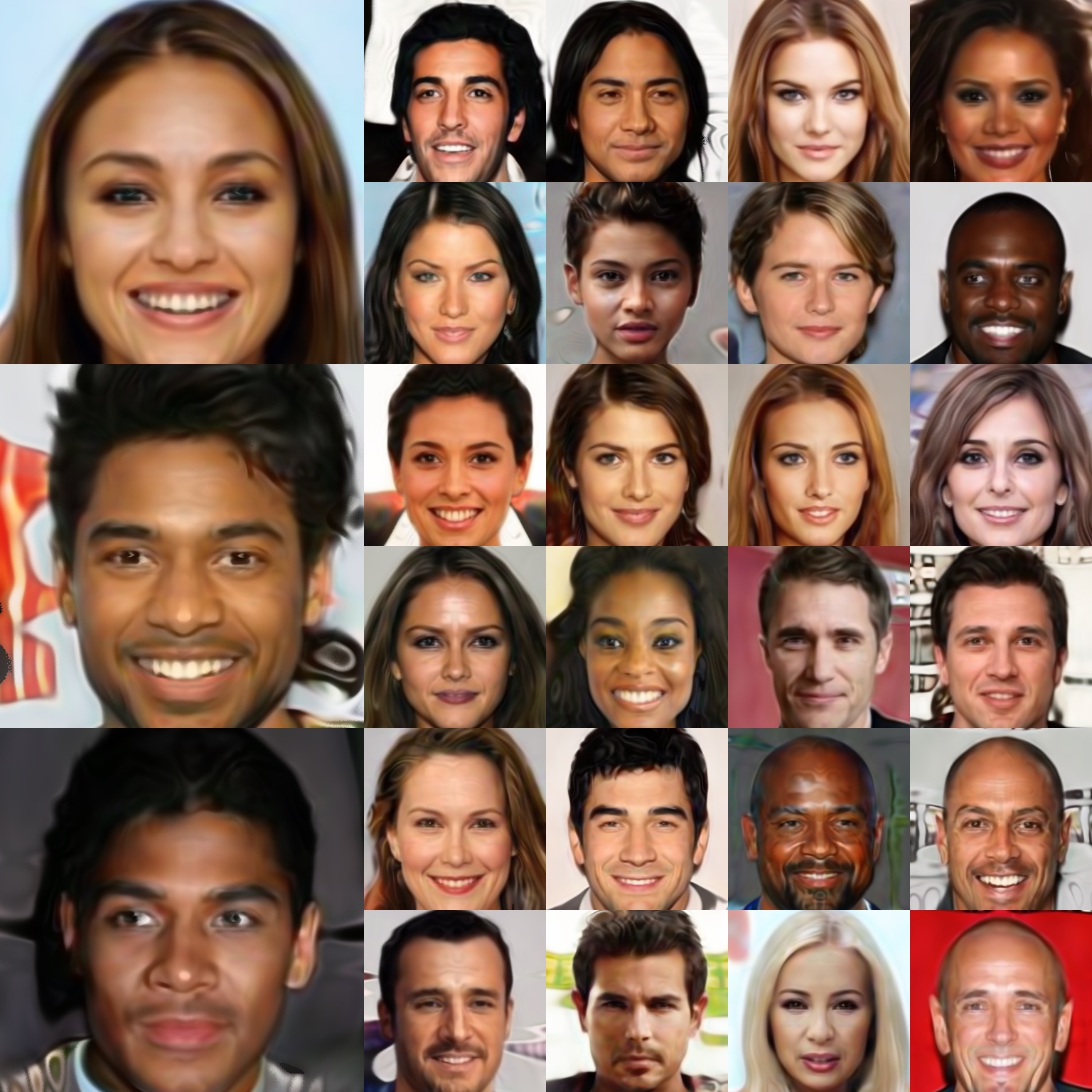}
    \caption{Curated examples from our model trained with CelebA~\cite{liu2015faceattributes}.}
    \label{fig:curated_celeba}
\end{figure*}

\begin{figure*}
    \centering
    \includegraphics[width=\textwidth]{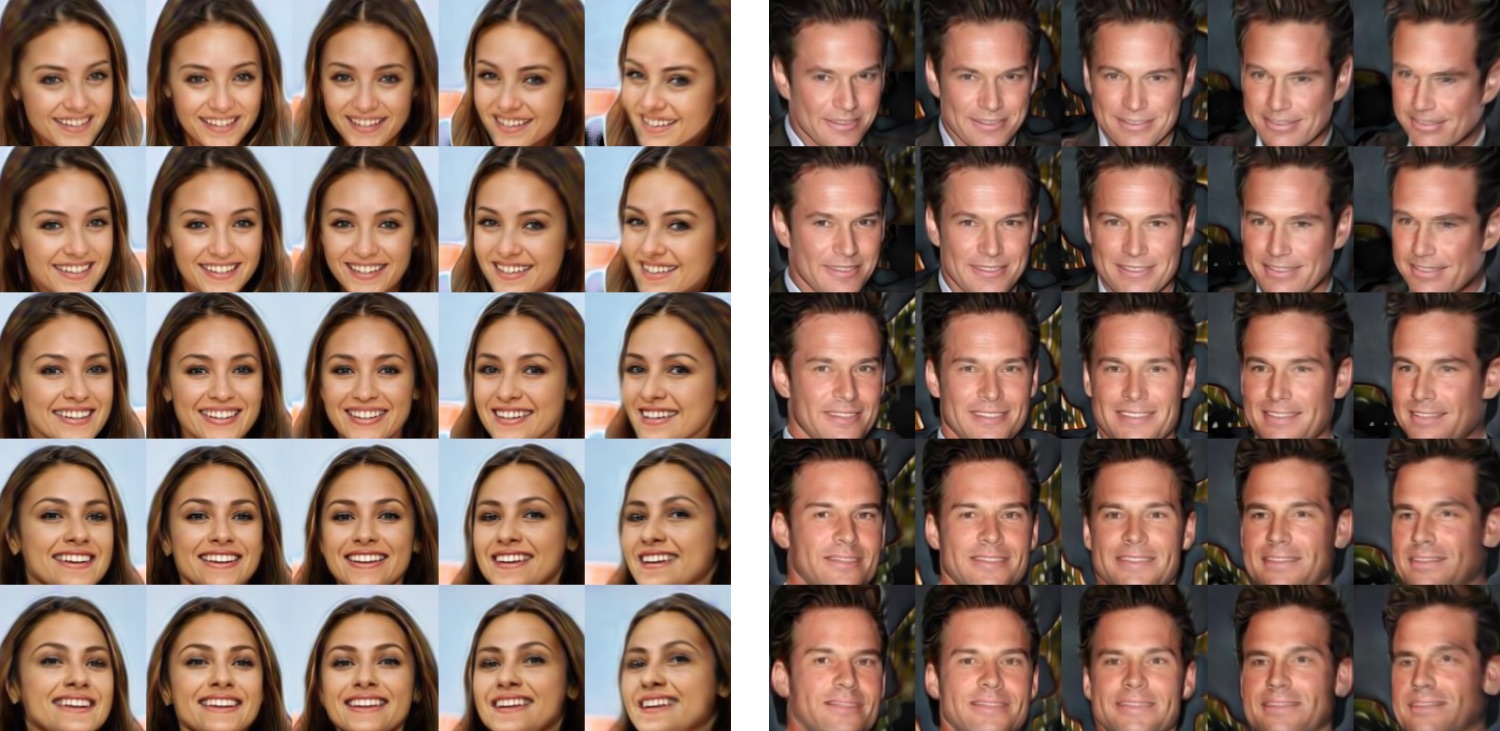}
    \caption{Curated examples from our model trained with CelebA, displayed from multiple viewing angles.}
    \label{fig:faces_multi_view}
\end{figure*}

\begin{figure*}
    \centering
    \includegraphics[width=\textwidth]{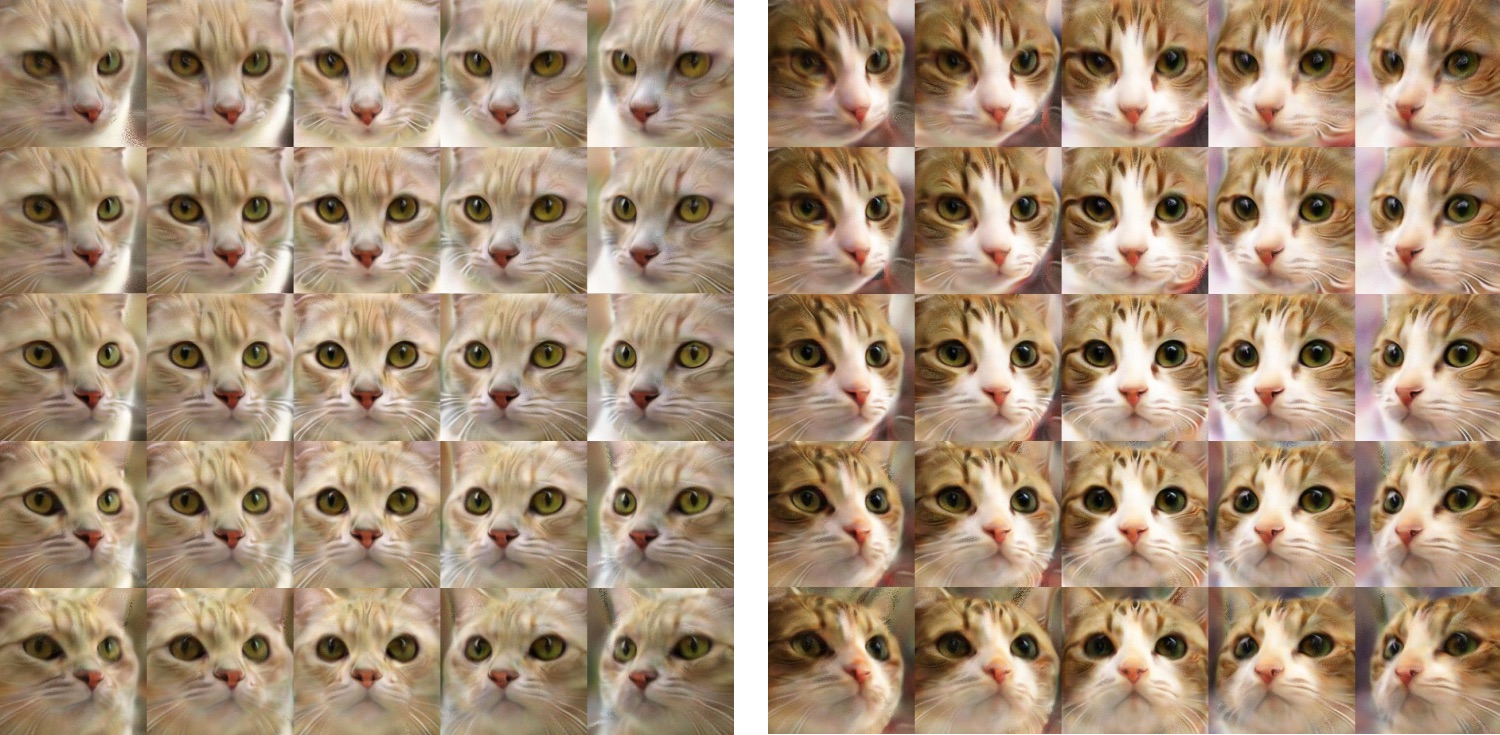}
    \caption{Curated examples from our model trained with Cats~\cite{cats}, displayed from multiple viewing angles.}
    \label{fig:cats_multi_view}
\end{figure*}

\begin{figure*}
    \centering
    \includegraphics[width=\textwidth]{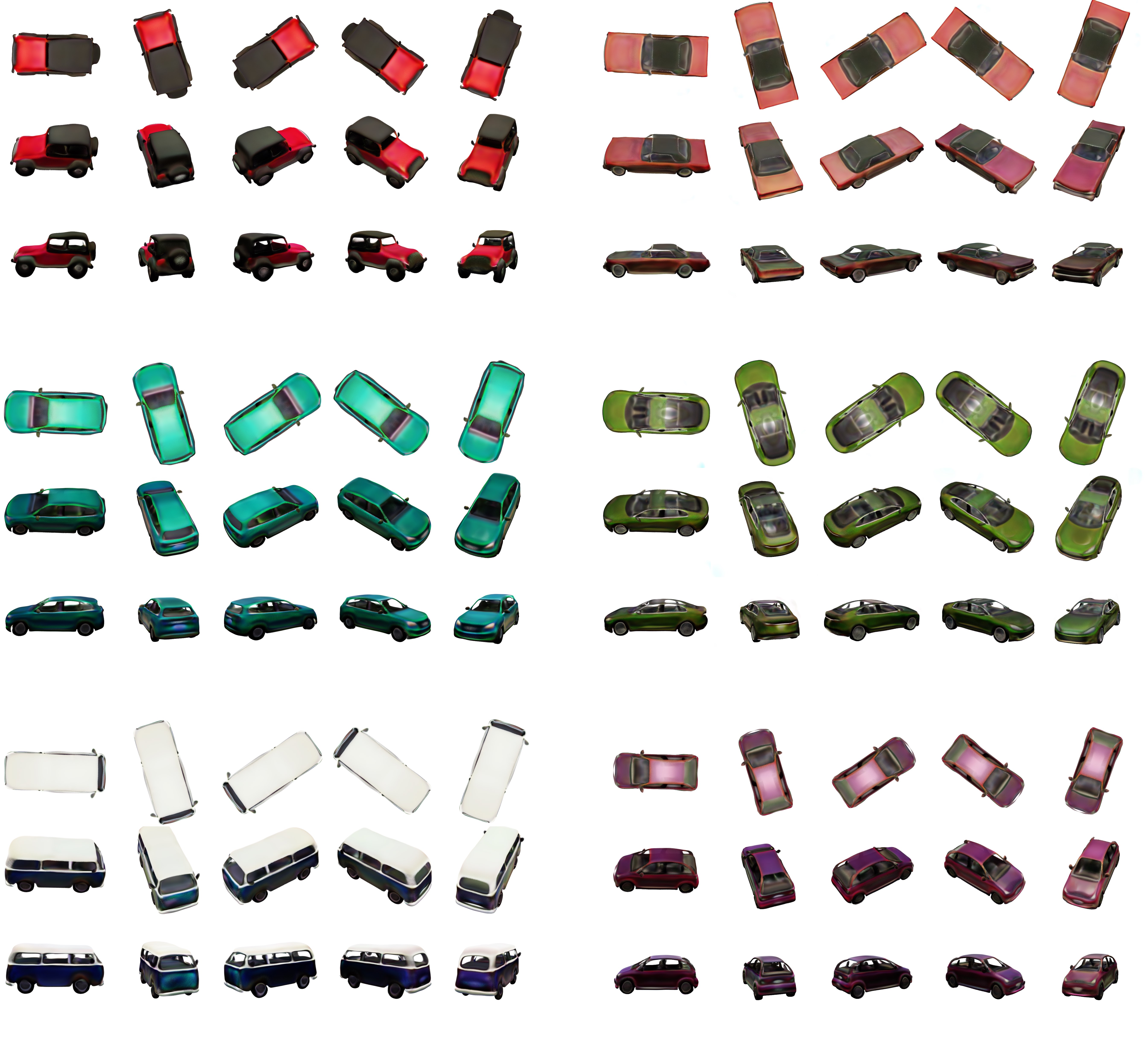}
    \caption{Curated examples from our model trained with CARLA~\cite{DBLP:journals/corr/abs-1711-03938}, displayed from multiple viewing angles.}
    \label{fig:cars_multi_views}
\end{figure*}

We include additional visual results to show the image quality and view consistency of \model. Figures~\ref{fig:faces_multi_view} and~\ref{fig:cats_multi_view} demonstrate the wide range of camera poses supported by \model for generated faces and cats. Figure~\ref{fig:curated_celeba} shows the fine detail that \model renders on larger images. Figure~\ref{fig:cars_multi_views} shows additional cars with varying elevation and rotation. We include several videos of faces and cats with the camera following an elliptical trajectory in our supplementary video.  

\section{COLMAP Reconstruction}
In order to demonstrate the images from \model are multi-view consistent, we include a COLMAP reconstruction in Figure~\ref{fig:COLMAP}. We observe that proxy shapes extracted from pi-GAN lead to more pleasing novel views when projected to novel camera poses than those from GRAF.

\section{Interpolation and Truncation}
Following the method of StyleGAN~\cite{karras2019style} we can smoothly interpolate between two generated samples by linearly interpolating between the frequencies and phase shifts corresponding to the two latent codes. We include a result in Figure~8 in  the paper. Along similar lines, it is also possible to trade off fidelity and diversity at test time following the method proposed in StyleGAN~\cite{karras2019style}. Because truncation reduced the diversity of generated images, we provided all evaluation metrics without truncation.

\section{Precision and Recall}
Recent work in generative models have investigated alternative metrics in order to independently evaluate fidelity and diversity~\cite{precision_recall_distributions, Kynkaanniemi2019}. Figure~\ref{fig:Precision-Recall} provides precision-recall plots on CelebA, Cats, and CARLA, comparing \model to GRAF and HoloGAN.

\end{document}